\documentclass{article}
\usepackage{graphicx}
\usepackage{amsmath,amssymb,amsfonts}%
\usepackage{times}
\usepackage{tabularx} 
\usepackage[title]{appendix}%
\usepackage{booktabs}%
\usepackage{multirow}%
\usepackage{hyperref}
\usepackage[a4paper, left=1.5in, right=1.5in, top=1in, bottom=1in]{geometry}
\title{Self-Supervised Learning for Neural Topic Models with Variance-Invariance-Covariance Regularization}
\author{
    Weiran Xu, Kengo Hirami, Koji Eguchi \\
    Graduate School of Advanced Science and Engineering, Hiroshima University \\
    \texttt{\{d210104, m231182\}@hiroshima-u.ac.jp}, \texttt{eguchi@acm.org}
}
\date{}
\begin{document}
\maketitle
\begin{abstract}
    In our study, we propose a self-supervised neural topic model (NTM) that combines the power of NTMs and regularized self-supervised learning methods to improve performance. 
    NTMs use neural networks to learn latent topics hidden behind the words in documents, enabling greater flexibility and the ability to estimate more coherent topics compared to traditional topic models. 
    On the other hand, some self-supervised learning methods use a joint embedding architecture with two identical networks that produce similar representations for two augmented versions of the same input. 
    Regularizations are applied to these representations to prevent collapse, which would otherwise result in the networks outputting constant or redundant representations for all inputs.
    Our model enhances topic quality by explicitly regularizing latent topic representations of anchor and positive samples. 
    We also introduced an adversarial data augmentation method to replace the heuristic sampling method. 
    We further developed several variation models including those on the basis of an NTM that incorporates contrastive learning with both positive and negative samples. 
    Experimental results on three datasets showed that our models outperformed baselines and state-of-the-art models both quantitatively and qualitatively.
\end{abstract}

\section{Introduction}\label{sec1}
Topic modeling is a statistical method to analyze large document collections. It extracts useful information called topics from document collections and the topics can be used for various downstream applications such as retrieval, summarization, sentiment analysis, etc~\cite{blei2012probabilistic,ijcai2021p638}.
The most representative topic model is latent Dirichlet allocation (LDA)~\cite{blei2003latent}, which uses a hierarchical Bayesian structure to learn latent topics assumed to be hidden behind the words in each document.  
With the development of deep neural networks and deep generative models, typically by adopting the variational auto-encoder (VAE)~\cite{welling2014auto} framework, more flexible neural topic models (NTMs) are being studied to handle large corpora, boost the performance of topic models, and potentially augment the performance in downstream applications without the need to design the inference process for conventional topic models.
Several variations of NTMs have been proposed, including those that replace the Dirichlet prior with other distributions~\cite{miao2017discovering,srivastava2017autoencoding}, those that simultaneously learn topics and word embeddings~\cite{dieng2020topic}, and those that incorporate external information~\cite{card2018neural,bai2018neural,hoyle2020improving,wang2021layer,wang2021extracting,tang2024beyond}.
In addition, NTMs tend to produce more coherent topics compared to LDA and achieve higher accuracy than supervised LDA~\cite{mcauliffe2007supervised} in document classification tasks. 
Models leveraging contrastive learning have also been proposed to further enhance NTM performance~\cite{nguyen2021contrastive,wu2022mitigating,zhang2022meta,han2023unified}.
Among these, Nguyen et al.~\cite{nguyen2021contrastive} introduced a contrastive learning NTM (CLNTM), which generates positive and negative samples based on the term frequency-inverse document frequency (tf-idf) of each word in each given document. This model improves performance by aligning the topic distribution of each anchor document closer to that of its positive sample while pushing it further from the negative sample. 
However, CLNTM faces an issue of negative samples may becoming similar to anchor samples as training progresses. As a result, the negative samples may no longer provide sufficient contrastive signals to effectively guide the learning process. 

In this study, we aimed to train an NTM with regularizations using a self-supervised learning approach that does not require negative samples. 
Self-supervised learning is valuable when true labels are unavailable in the training data or when annotation costs are high. 
In computer vision, self-supervised learning models have achieved performance comparable with supervised models. 
One approach that has attracted attention involves using two networks that share parameters and generate similar image representations for a pair of transformed images derived from the same source image.
However, this joint embedding architecture has a collapse problem, where the networks constantly output identical embeddings (referred to as `representation collapse') or redundant representations for all inputs (referred to as `dimension collapse'\footnote{In the context of NTMs, this `redundant representations for all inputs’ is also referred to as `component collapse’ or `topic collapse’~\cite{srivastava2017autoencoding,wu2023effective}.}).
To prevent such a problem, the intuitive and straightforward self-supervised learning method Variance-Invariance-Covariance Regularization (VICReg)~\cite{bardes2022vicreg} was proposed and applied to image data. 
VICReg maximizes the information content of embeddings by imposing specific restrictions so that the variables can carry unique information of the data.
This simple method achieves an image classification accuracy on par with state-of-the-art supervised methods by minimizing the distance between two embeddings from the same image, ensuring the embeddings for different images to be different, and decorrelating the variables of each embedding to prevent the dimension collapse.

To enhance topic quality, we drew inspiration from CLNTM and intuitively applied a self-supervised learning method to an NTM, focusing on explicitly regularizing latent representation of each document.
To the best of our knowledge, this study is the first to train an NTM with regularizations in a self-supervised learning manner.
Inspired by CLNTM, we propose a self-supervised NTM with regularizations, \textbf{VICNTM}, using the anchor and positive samples in the NTM's objective, which we extend using SCHOLAR~\cite{card2018neural}, as previously done by Nguyen et al.~\cite{nguyen2021contrastive}, and also using the ideas inspired by VICReg~\cite{bardes2022vicreg}.
This approach preserves distinctions between latent topic representations of different documents, inherently playing the role of negative samples, while minimizing the linear correlation between representation dimensions to maximize informational content of representations.
Together, these two key regularizations on anchor and positive samples work to improve overall topic quality. 
In addition, our approach avoids the issue of negative samples being similar to the input, as seen in CLNTM, by eliminating the need for negative samples altogether.
Moreover, the explicit regularizations are capable of addressing component collapse or topic collapse, where all topics become identical during inference with a VAE---a challenge also addressed in some previous studies~\cite{srivastava2017autoencoding,wu2023effective}.
Inspired by Suzuki~\cite{Suzuki_2022_CVPR} in the field of computer vision, we also introduce a new data augmentation method based on an adversarial strategy to generate positive text samples.
We conducted experiments on three different datasets to evaluate the performance of our proposed model.
Moreover, we developed several variants of our model for comparison. 
The results indicate that our models outperformed two baselines and state-of-the-art models in terms of topic coherence, both quantitatively and qualitatively. 
The contributions of this work are summarized as follows:
\begin{enumerate}
    \item We developed a self-supervised NTM with regularizations and its variants where anchor samples and positive samples are explicitly regularized to produce better topics. To the best of our knowledge, this is the first study to learn NTMs with regularization in a self-supervised manner.
    \item We imposed three regularizations inspired by VICReg that was proposed in the image domain to the text samples in the latent topic space to improve topic quality. Two key regularizations are particularly important: one preserves differences between representations, and the other maximizes information content of the representations.
    \item To generate positive samples, we also introduced a model-based adversarial data augmentation method, replacing the heuristic tf-idf-based strategy.
    \item Our models outperformed two baselines and state-of-the-art models in terms of topic coherence, perplexity, and topic diversity on three different datasets. We also provided examples of topics and visualizations of generated latent topic representations to qualitatively demonstrate the superiority of our generated topics over those of the comparison methods.
\end{enumerate}

\section{Related work}\label{sec2}

\subsection{Neural Topic Models}
Topic modeling is a probabilistic method that maps a large collection of documents into a low-dimensional latent topic space.
LDA~\cite{blei2003latent}, the prominent topic model learned using a Bayesian inference process, generates each word of each document from a topic-word distribution in accordance with its latent topic specified by a document-topic distribution sampled from a Dirichlet prior. 
NTMs, mainly on the basis of VAE~\cite{welling2014auto}, approximate the posterior distribution of topics using an encoder in the inference process and generate a bag-of-words (BoW) representation of each docuement using a decoder. 
ProdLDA~\cite{srivastava2017autoencoding} is the first to introduce a logistic normal prior into a VAE-based NTM to approximate a Dirichlet prior as in LDA. 
To mitigate the issue of topic collapsing, which commonly occurs in VAE-based models, Wu et al.~\cite{wu2023effective} proposed a VAE-based NTM with embedding clustering regularization. This regularization sets the topic embeddings as the cluster centers of the word embeddings and ensures that the clusters are as far apart as possible.
Building upon ProdLDA, SCHOLAR~\cite{card2018neural} is a general NTM that incorporates various types of metadata and utilizes the background log-frequency of words. It achieves better topic quality since it excludes common words across different topics. It can be trained in both supervised and unsupervised settings.

Apart from SCHOLAR, researchers developed NTMs that incorporate external information to enhance topic quality.
Bai et al.~\cite{bai2018neural} incorporate an external relational citation network of documents into NTMs, where the topic distributions of two documents are fed into a neural network with multilayer perceptrons (MLPs) to predict whether they should be connected.
Wang et al.~\cite{wang2021layer} jointly encode texts and their network links to derive topic distributions using an augmented encoder network, which consists of an MLP and a graph convolutional network.
To address the sparsity problem of short texts, Wang et al.~\cite{wang2021extracting} enrich the BoW representations by incorporating a word co-occurrence graph and constructing a word semantic correlation graph using pre-trained word embeddings.
On the basis of SCHOLAR, SCHOLAR+BAT~\cite{hoyle2020improving} leverages the rich knowledge from a pre-trained BERT~\cite{devlin2019bert} to guide the learning of the NTM in a knowledge distillation framework.
In a more recent study, Tang et al.~\cite{tang2024beyond} incorporate a structural causal module into an NTM to simultaneously capture causal relationships between topics and metadata, as well as relationships within the elements themselves.
In this study, we concentrate on NTMs that do not utilize external information. 
However, our models can be easily extended to incorporate external information, which could potentially lead to further performance enhancements.

\subsection{Neural Topic Models with contrastive learning}
In this field, research typically begin by identifying positive and negative samples at either the document level or latent topic distribution level. Then, contrastive learning is applied to the topic distribution level alongside the standard NTM learning objectives. 
Also on the basis of SCHOLAR, CLNTM~\cite{nguyen2021contrastive} uses a data augmentation method for texts that uses the tf-idf of each word in a document to create positive and negative samples. Positive (negative) samples are generated by replacing the values of insignificant (salient) words in the input BoW vector with the values of those chosen words in the recostructed BoW vector. This enables the application of contrastive learning, where the anchor sample's prototype representation is moved closer to the representation of the positive sample and pushed further from the representation of the negative sample.
Han et al.~\cite{han2023unified} combine clustering based on a pre-trained language model with an NTM. They use contrastive learning to cluster document embeddings and select salient words from each cluster to form the vocabulary for the NTM. Contrastive learning is then applied again during the NTM training stage by leveraging the topic distributions of positive samples and the prior distribution.
For short texts, Wu et al.~\cite{wu2022mitigating} select positive and negative samples based on topic distributions to mitigate the data sparsity problem. 
Zhang et al.~\cite{zhang2022meta} improve short text topic modeling in variable-length corpora by transferring semantic information from long texts to short texts, learning the semantic relationships through a contrastive learning approach.

\subsection{Self-supervised learning}
To address the collapse problem in joint embedding architectures, several intuitive self-supervised learning methods have been proposed to regularize the content of the embeddings.
Barlow Twins~\cite{zbontar2021barlow} makes the normalized cross-correlation matrix between two embeddings of distorted versions of a sample move toward the identity matrix.
VICReg~\cite{bardes2022vicreg} minimizes the mean square distance between the two embeddings, ensures the standard deviation of each element remains above a threshold, and reduces the covariances between each pair of embedding variables toward zero.
Both methods project the representation vectors into a high-dimensional space where the regularization on the embeddings indirectly reduces the redundant information in the representation vectors.

\section{Methodology}\label{sec3}
\begin{figure*}[ht]
    \centering
    \includegraphics[width=1.0\textwidth]{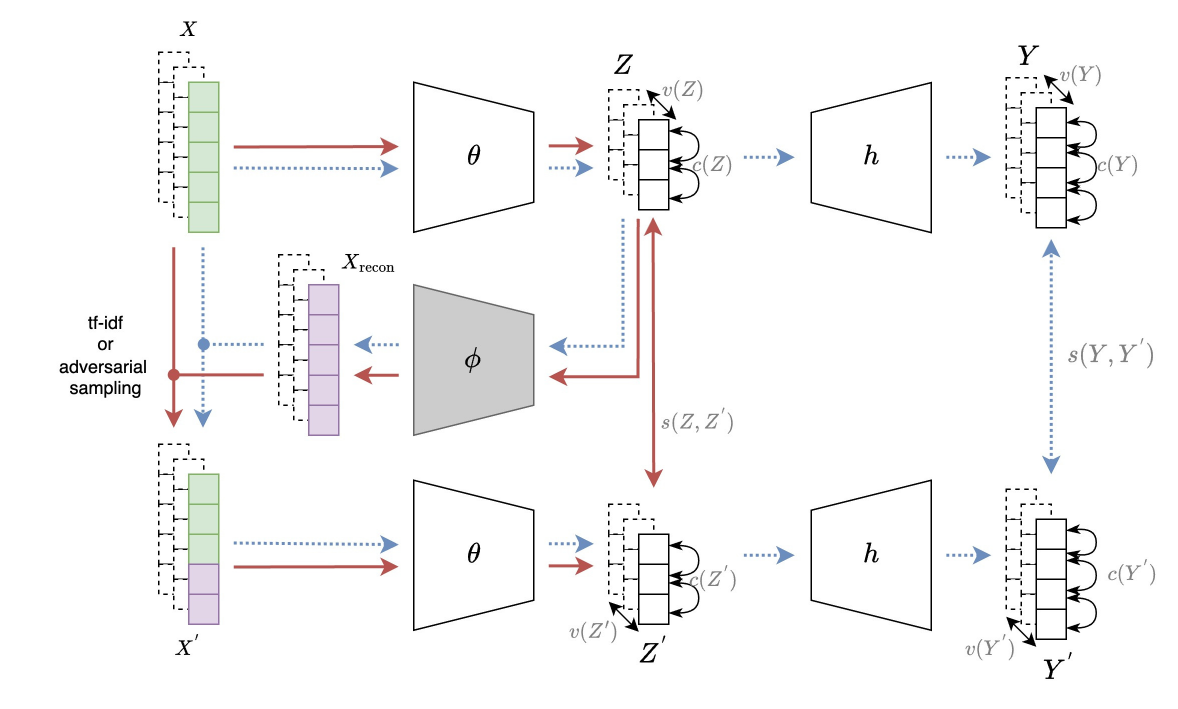}
    \caption{{\bf Illustration of VICNTM and Deep VICNTM.} \textbf{\textit{VICNTM}}: components connected by red solid arrows. \textbf{\textit{Deep VICNTM}}: components connected by blue dot arrows. Note that VICNTM performs regularization on the latent representation $\boldsymbol{Z}$ and $\boldsymbol{Z'}$, while Deep VICNTM performs regularization on the high-dimensional embeddings $\boldsymbol{Y}$ and $\boldsymbol{Y'}$} \label{fig:vicntm}
\end{figure*}

Inspired by CLNTM~\cite{nguyen2021contrastive}, we develop our model by applying regularization inspired by VICReg~\cite{bardes2022vicreg} to SCHOLAR~\cite{card2018neural}.
In this section, we will begin by providing a concise overview of VICReg, SCHOLAR and strategies for sampling positive texts in Section~\ref{vicreg}, \ref{scholar} and~\ref{sampling strategies}, respectively. 
These elements form the fundamental basis for our models. 
Moving on to Section~\ref{vicntm}, we will present a detailed description of our proposed model, VICNTM, as well as introduce a variation model, Deep VICNTM.
Both models are based on SCHOLAR in the unsupervised setting, with the major distinction being that VICNTM applies the VIC regularization in the latent topic space, whereas Deep VICNTM performs the regularization in a high-dimensional space, similar to VICReg, as illustrated in Fig.~\ref{fig:vicntm}.

\subsection{Preliminary: VICReg}\label{vicreg}
\begin{figure*}[ht]
    \centering
    \includegraphics[width=1.0\textwidth]{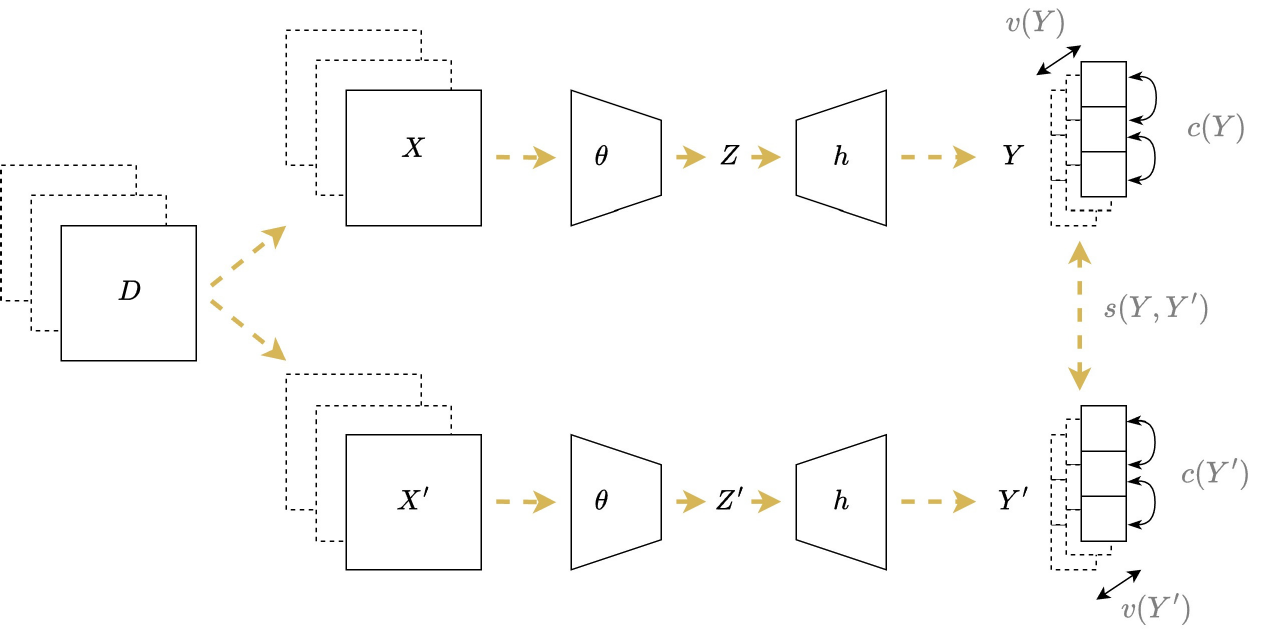}
    \caption{{\bf Illustration of VICReg.} VICReg performs regularization on the high-dimensional embeddings $\boldsymbol{Y}$ and $\boldsymbol{Y'}$} \label{fig:vicreg}
\end{figure*}

As illustrated in Fig.~\ref{fig:vicreg}, the self-supervised learning method VICReg has two branches, each of which inputs the distorted version $\boldsymbol{X}$ or $\boldsymbol{X'}$ of the minibatch $D$ composed of $n$ data vectors, outputs $n \times k$-dimensional representation vectors $\boldsymbol{Z}$ or $\boldsymbol{Z'}$ in a latent space via the encoder $\theta$, and then via the expander $h$ outputs $n \times d$-dimensional embedding vectors $\boldsymbol{Y}$ or $\boldsymbol{Y'}$ in a high-dimensional projection space where the regularizations are executed in accordance with the following objective:
\begin{eqnarray} \label{eq:vicreg}
    \mathcal{L}_{\text{VICReg}}(\boldsymbol{Y},\boldsymbol{Y'}) = \mu \underbrace{[v(\boldsymbol{Y})+v(\boldsymbol{Y'})]}_{\boldsymbol{V}ariance} + \lambda \underbrace{s(\boldsymbol{Y},\boldsymbol{Y'})}_{\boldsymbol{I}nvariance} +\nu \underbrace{[c(\boldsymbol{Y})+c(\boldsymbol{Y'})]}_{\boldsymbol{C}ovariance}, 
\end{eqnarray}
where
\begin{eqnarray*}
    v(\boldsymbol{Y}) =& \displaystyle\frac{1}{d}\displaystyle\sum_{j=1}^{d}\max(0,\gamma-\sqrt{\text{Var}(\boldsymbol{Y}^j)+\epsilon}), \notag \\
    s(\boldsymbol{Y},\boldsymbol{Y'})  =& \displaystyle\frac{1}{n}\displaystyle\sum_{i}\|\mathbf{y}_i-\mathbf{y}_i^{'}\|_2^2, \notag \\
    c(\boldsymbol{Y}) =& \displaystyle\frac{1}{d} \displaystyle\sum_{i \neq j}[C(\boldsymbol{Y})]_{i,j}^2, \notag \\
    C(\boldsymbol{Y}) =& \displaystyle\frac{1}{n-1}\displaystyle\sum_{i=1}^{n}(\mathbf{y}_i-\bar{\mathbf{y}})(\mathbf{y}_i-\bar{\mathbf{y}})^T, \bar{\mathbf{y}}=\displaystyle\frac{1}{n}\displaystyle\sum_{i=1}^{n}\mathbf{y}_i.
\end{eqnarray*}
The three terms $s(\boldsymbol{Y},\boldsymbol{Y'})$, $v(\boldsymbol{Y})$ and $c(\boldsymbol{Y})$ in Eq.(\ref{eq:vicreg}) correspond to the \textit{invariance} that captures the relation between the two branches, the \textit{variance} that captures the relation of embeddings within a minibatch, and the \textit{covariance} that captures the relation between each pair of dimensions of embeddings within a minibatch, respectively. 
Minimizing the three terms ensures that the embeddings from two minibatches ($\boldsymbol{Y}$ and $\boldsymbol{Y'}$) are as similar as possible, the embeddings within a minibatch ($\boldsymbol{Y}$ or $\boldsymbol{Y'}$) are diverse from each other, and the dimensions of the embeddings within a minibatch ($\boldsymbol{Y}$ or $\boldsymbol{Y'}$) are decorrelated from each other.
As a result, this process indirectly minimizes the differences between $\boldsymbol{Z}$ and $\boldsymbol{Z'}$, preserving the  distinctions between different latent representations while reducing dependencies between dimensions of latent representation space.
These representations are then utilized for downstream tasks after the training process.
The notations $\lambda$, $\mu$, and $\nu$ are hyperparameters that control the balance among loss terms in Eq.(\ref{eq:vicreg}).

\subsection{SCHOLAR}\label{scholar}
SCHOLAR is an NTM based on the VAE framework that incorporates external variables for generalization.
We focused only on the unsupervised part of this model, which can also be seen as ProdLDA~\cite{srivastava2017autoencoding} with the background log-frequency of words.
This model enhances performance by replacing topics with sparse deviations from the background log-frequency.
For each BoW representation $\mathbf{x} \in \mathbb{R}^{\mathcal{|V|}}$ and its document-topic distribution $\mathbf{z} \in \mathbb{R}^k$ where $|\mathcal{V}|$ is the number of word types and $k$ is the number of topics, the model is learned by optimizing an encoder parameterized by 
$\boldsymbol{\theta} = \{\boldsymbol{\theta}_1, \boldsymbol{\theta}_2\}$ and a decoder parameterized by $\boldsymbol{\phi}=\{\boldsymbol{\beta}\}$, where $\boldsymbol{\beta}$ is the topic-word distribution matrix for all topics.
Specifically, the neural variational inference is conducted as follows. The encoder first employs MLPs to compute $\boldsymbol{\mu}=f_{\theta_1}(\mathbf{x})$ and $\boldsymbol{\Sigma}=\text{diag}(f_{\theta_2}(\mathbf{x}))$, where $\text{diag}(\cdot)$ represents the conversion of a vector into a diagonal matrix. 
Then $\boldsymbol{r}$ is sampled by the reparameterization trick~\cite{welling2014auto}: $\boldsymbol{r}=\boldsymbol{\mu}+(\boldsymbol{\Sigma})^{\frac{1}{2}}\boldsymbol{\epsilon}$, where $\boldsymbol{\epsilon}\sim \mathcal{N}(\boldsymbol{0}, \boldsymbol{\mathrm{I}})$.
After that, $\mathbf{z}$ is modeled as $\mathbf{z}=\text{softmax}(\boldsymbol{r})$.
The decoder then reconstructs each BoW representation $\mathbf{x'}\sim \text{Mult}(\text{softmax}(\boldsymbol{d}+\mathbf{z}^{\top}\boldsymbol{\beta}))$, where $\boldsymbol{d}$ is the background log-frequency.
The model is learned by minimizing the following objective:
\begin{eqnarray} \label{eq:ntm}
    \mathcal{L}_\text{NTM} = -\mathbb{E}_{q_{\boldsymbol{\theta}}(\mathbf{z}|\mathbf{x})}[\log p_{\boldsymbol{\phi}}(\mathbf{x}|\mathbf{z})]+\mathbb{KL}[q_{\boldsymbol{\theta}}(\mathbf{z}|\mathbf{x})\|p(\mathbf{z})].
\end{eqnarray}
The first term minimizes the negative expected log-likelihood $p(\mathbf{x}|\mathbf{z})$ on the variational distribution $q(\mathbf{z}|\mathbf{x})$, which is computed as $-\mathbf{x}^{\top}\log (\mathbf{x'})$.
The second term minimizes the Kullback-Leibler (KL) divergence between the posterior and prior, which is a logistic normal one.

\subsection{Sampling strategies}\label{sampling strategies}
When incorporating self-supervised learning into the originally unsupervised NTMs, we use the sampling method proposed by Nguyen et al.~\cite{nguyen2021contrastive} and explore a new data augmentation method based on an adversarial strategy, additionally.
\par
\noindent
{\bf Tf-idf-based strategy} is on the basis of the assumption that people can distinguish the similarity between two documents if they have a similar proportion of similar salient words. 
As illustrated in Fig.~\ref{fig:vicntm}, we sample the positive samples $\boldsymbol{X'}$ by replacing the values of the $t$ unimportant tokens that have the lowest tf-idf values in the input BoW vectors $\boldsymbol{X}$ with the values of those unimportant tokens in the reconstructed BoW vectors $\boldsymbol{X}_{\text{recon}}$ generated by the decoder in SCHOLAR.
\par
\noindent
{\bf Adversarial strategy} is inspired by TeachAugment~\cite{Suzuki_2022_CVPR}, a data augmentation model that alternates between learning an augmentation model and a target model leveraging knowledge from an EMA teacher model\footnote{The weights of the teacher model are updated using an exponential moving average (EMA) of the weights from the target model.} in an adversarial manner.
Within the context of image classification tasks, TeachAugment generates augmented images that are adversarial for the target classification model but recognizable for the teacher classification model.
Specifically, the learning procedure is as follows:
\begin{enumerate}
    \item Update the target model while keeping the augmentation model fixed
    \begin{itemize}
        \item Minimize loss for the target model
    \end{itemize}
    \item Update the augmentation model while keeping the target model fixed
    \begin{itemize}
        \item Maximize loss for the target model
        \item Minimize loss for the teacher model
    \end{itemize}
    \item Iterate the steps outlined above
\end{enumerate}
This adversarial strategy ensures the generated positive samples are recognizable for the teacher model, such that they can be different from but still similar to the anchor samples. 
Analogous to the original TeachAugment, in this paper, we utilize document indices as class labels to ensure that the generated positive samples (at the document level) are correctly classified back to the anchor samples from which they are augmented. 
We simply implement the augmentation model in TeachAugment as $\mathbf{x}_{\text{aug}}=\mathbf{x}+g(\mathbf{x})$, where $g(\mathbf{x})$ models the difference between the original input $\mathbf{x}$ and the augmented input $\mathbf{x}_{\text{aug}}$. $g(\mathbf{x})$ is a transformation of $\mathbf{x}$ through a linear layer followed by ReLU activation. Both the target model and the teacher model in TeachAugment are implemented as MLP classifiers.

The tf-idf-based strategy generates positive samples during the training of NTM, which leads to the issue of these samples becoming nearly identical to the anchor samples as the training progresses. 
The adversarial strategy does not face this problem as it generates samples prior to the training phase. 
However, we primarily use the former strategy in the following experiments, and additionally, we employ the latter one.

\subsection{VICNTM}\label{vicntm}

By integrating SCHOLAR with regularization inspired by VICReg, we propose a self-supervised neural topic model \textbf{VICNTM} that incorporates variance-invariance-covariance (VIC) regularization into the latent topic representations.
The proposed model is illustrated in Fig.~\ref{fig:vicntm}. 
The model first inputs the BoW representations of anchor documents $X_{\text{BoW}}$ and generates positive samples $X'_{\text{BoW}}$ in accordance with the tf-idf value of each token.  
The model then executes VIC regularization on the latent topic representations $\boldsymbol{Z}$ and $\boldsymbol{Z'}$. 
Note that the original VICReg performs regularization in the high-dimensional projection space, while our model performs regularization in the topic space.
The topics learned with our model are expected to be diverse with less redundant information by maintaining the differences between different latent topic representations of documents and minimizing the linear correlation between different topics.
Given a minibatch of $N$ documents, our model is learned by minimizing the combination of Eq.(\ref{eq:ntm}) and Eq.(\ref{eq:vicreg}), with Eq.(\ref{eq:vicreg}) specifically applied to the latent representations:
\begin{eqnarray} \label{eq:vicntm}
    \mathcal{L} =& \mathcal{L}_{\text{NTM}} + \mathcal{L}_{\text{VICReg}}(\boldsymbol{Z},\boldsymbol{Z'}) \notag \\
    =& \left( \sum_{i}^{N}-\mathbb{E}_{q_{\theta}(\mathbf{z}_i|\mathbf{x}_i)}[\log p_{\phi}(\mathbf{x}_i|\mathbf{z}_i)]+\mathbb{KL}[q_\theta(\mathbf{z}_i|\mathbf{x}_i)\|p(\mathbf{z}_i)]\right) \notag \\
    +& \lambda s(\boldsymbol{Z},\boldsymbol{Z'}) +\mu [v(\boldsymbol{Z})+v(\boldsymbol{Z'})] \notag \\
    +& \nu [c(\boldsymbol{Z})+c(\boldsymbol{Z'})]
\end{eqnarray}

In addition, we propose a variation of VICNTM, \textbf{Deep VICNTM}, as illustrated in Fig.~\ref{fig:vicntm}.
Deep VICNTM performs VIC regularization in high-dimensional space where embeddings $\boldsymbol{Y}$ and $\boldsymbol{Y'}$ are projected from latent topic representations $\boldsymbol{Z}$ and $\boldsymbol{Z'}$ respectively, using an expander $h$ as in VICReg.
The training objective needs to be modified by simply replacing $\boldsymbol{Z}$ and $\boldsymbol{Z'}$ with $\boldsymbol{Y}$ and $\boldsymbol{Y'}$ respectively, as follows:
\begin{eqnarray} \label{eq:deepvicntm}
    \mathcal{L} =& \mathcal{L}_{\text{NTM}} + \mathcal{L}_{\text{VICReg}}(\boldsymbol{Y},\boldsymbol{Y'}) \notag \\
    =& \left( \sum_{i}^{N} -\mathbb{E}_{q_{\theta}(\mathbf{z}_i|\mathbf{x}_i)}[\log p_{\phi}(\mathbf{x}_i|\mathbf{z}_i)]+\mathbb{KL}[q_\theta(\mathbf{z}_i|\mathbf{x}_i)\|p(\mathbf{z}_i)]\right) \notag \\
    +& \lambda s(\boldsymbol{Y},\boldsymbol{Y'}) +\mu [v(\boldsymbol{Y})+v(\boldsymbol{Y'})] \notag \\
    +& \nu [c(\boldsymbol{Y})+c(\boldsymbol{Y'})],
\end{eqnarray}
where
\begin{eqnarray*}
    \boldsymbol{Y} =& h(\boldsymbol{Z}), \\ \notag
    \boldsymbol{Y'} =& h(\boldsymbol{Z'}).
\end{eqnarray*}

\section{Experiments}\label{sec4}
In this section, we begin by introducing the datasets and baselines in Section~\ref{sec41}, followed by the evaluation metrics adopted, which are detailed in Section~\ref{sec42}. 
Regarding the two sampling strategies mentioned earlier in Section~\ref{sampling strategies}, the results from experiments utilizing the tf-idf-based sampling strategy and findings are elaborated from the model structure perspective in Section~\ref{sec43} and~\ref{sec44}. 
Furthermore, the results from experiments employing the adversarial sampling strategy are presented in Section~\ref{sec45}.

\subsection{Settings}\label{sec41}
We conducted the experiments on three datasets to evaluate the topic quality of our proposed model.
Each dataset was preprocessed to remove stopwords, words with only one character, words with a document frequency less than 100, and words with a document frequency greater than 70\% of the total number of documents. 
We filtered out documents with fewer than 30 word types to ensure the replacement when generating positive samples.
Table~\ref{tab:datasets} summarizes the datasets, where $|\mathcal{D}|$ and $|\mathcal{V}|$ indicate the number of documents and the number of word types, respectively.

\begin{table*}[ht]
    \caption{Dataset details}\label{tab:datasets}%
    \begin{tabular*}{\textwidth}{@{\extracolsep\fill}lccc}
    \toprule
    & $|\mathcal{D}|$   & $|\mathcal{V}|$   & train/dev/test split \\ \midrule
    20Newsgroups (20NG)~\cite{lang1995newsweeder}       & 13k & 3k  & 48/12/40             \\
    IMDb movie reviews (IMDb)~\cite{maas2011learning} & 43k & 5.7k  & 50/25/25             \\
    Wikitext-103 (Wiki)~\cite{merity2017pointer}       & 28.5k & 21k & 70/15/15             \\ \bottomrule
    \end{tabular*}
\end{table*}

We compared our model with the following models:
\begin{description}
\item[\rm{\textbf{ProdLDA}~\cite{srivastava2017autoencoding}:}] the most popular and conventional VAE-based NTM implemented with logistic normal prior, as a baseline.
\item[\rm{\textbf{ECRTM}~\cite{wu2023effective}:}] a state-of-the-art VAE based NTM with embedding clustering regularization. 
\item[\rm{\textbf{TSCTM}~\cite{wu2022mitigating}:}] a topic semantic contrastive NTM designed for short texts.
\item[\rm{\textbf{UTopic}~\cite{han2023unified}:}] an NTM that combines clustering and a VAE-based NTM, employing contrastive learning at both stage. 
\item[\rm{\textbf{SCHOLAR}~\cite{card2018neural}:}] a VAE-based general NTM implemented with logistic normal prior and background log-frequency as a baseline.
\item[\rm{\textbf{CLNTM}~\cite{nguyen2021contrastive}:}] a SCHOLAR-based NTM using positive and negative samples for contrastive learning as a state-of-the-art baseline.

\end{description}
Additionally, we developed three more variation models by applying an expander to high-dimensional embedding space and/or performing regularization to CLNTM:
\begin{description}
\item[\rm{\textbf{VC-CLNTM}}]performs variance-covariance (VC) regularization to latent topic representations of anchor and positive samples in CLNTM.
\item[\rm{\textbf{Deep VC-CLNTM}}]performs VC regularization in high-dimensional space where embeddings are projected from latent topic representations of anchor and positive samples in CLNTM, while the contrastive learning part remains in the latent topic space.
\item[\rm{\textbf{VIC-CLNTM}}]performs VIC regularization to latent topic representations of anchor and positive samples in CLNTM while the contrastive learning part is replaced with the cosine similarity between the latent topic representations of anchor and negative samples.
\end{description}

\subsection{Metrics}\label{sec42}
We conducted experiments with two different numbers of topics, $k=50$ and $k=200$, to evaluate the models' performance, following the previous studies~\cite{srivastava2017autoencoding,hoyle2020improving,nguyen2021contrastive}. 
Using different random seeds, we ran ten trials to measure topic coherence, perplexity, and topic diversity. 
For topic coherence, we used normalized pointwise mutual information (NPMI)~\cite{lau2014machine} of the top-ten words of each topic. 
For topic diversity, we used Topic Diversity (TD)~\cite{dieng2020topic} defined as the percentage of unique words in the top-ten words of all topics, and Inversed Rank-Biased Overlap (IRBO)~\cite{terragni2021word,bianchi2021pre}, which measures how diverse the combinations of all topics are with weighted ranking in the top-ten words of each topic. IRBO ranges from 0 when topics are identical to 1 when topics are entirely dissimilar.

\subsection{Implementation details}\label{sec43}
\begin{figure*}[ht]
    \centering
    \includegraphics[width=1.0\textwidth]{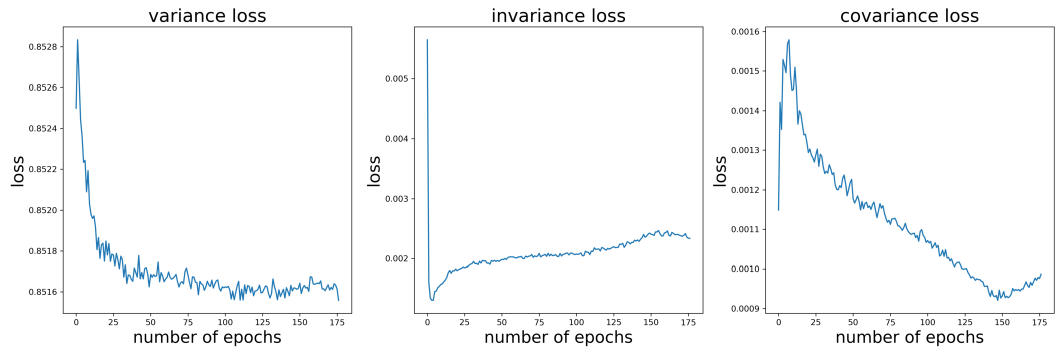}
    \caption{{\bf Learning curves that confirm the three regularization losses were effectively learned}} \label{fig:losscurves}
\end{figure*}

Each expander in our deep variation models is composed of three fully-connected layers with ReLU.
To determine the optimal hyperparameters $\lambda$, $\mu$, and $\nu$ for each of our proposed models' VIC regularization terms and each expander's output dimension, we used Bayesian optimization\footnote{In our experiments, we used 'Optuna'~\cite{akiba2019optuna}, a Python library that automates the process of hyperparameter tuning, which is available at https://optuna.org/.} on the basis of NPMI on the validation set when $k=50$. 
We use the same hyperparameters for the experiments when $k=200$ as those when $k=50$.
We also implemented early-stopping to stop training if the NPMI on the validation set did not increase for 30 epochs. 
The batch sizes for the experiments are 50, 1000, and 250 for the 20NG, IMDb, and Wiki datasets, respectively.
The learning curves in Fig.~\ref{fig:losscurves} represnt the result of one of our experiments for VICNTM, illustrating that the VIC regularization was effectively learned in our models, as an example.

\subsection{Results}\label{sec44}
\subsubsection{Quantitative Results}
\begin{table*}[ht]
\centering
\caption{Results of NPMI of NTMs when $k=50$. Boldface indicates the best performance for each experiment. Underlining indicates superior performance of our models compared to the baselines for each experiment}
\label{tab:results50}
\begin{tabular*}{\textwidth}{@{\extracolsep\fill}lccc}
    \hline
    $k=50$              & 20NG                   & IMDb                    & Wiki                    \\ \hline
    ProdLDA       & 0.2324{\tiny $\pm0.0093$}          & 0.0828{\tiny $\pm0.0033$}           & 0.2499{\tiny $\pm0.0117$}           \\
    ECRTM         & 0.2299{\tiny $\pm0.0127$}          & 0.0775{\tiny $\pm0.0144$}           & 0.3960{\tiny $\pm0.0194$}           \\
    TSCTM         & 0.2505{\tiny $\pm0.0215$}          & 0.0766{\tiny $\pm0.0192$}           & 0.3481{\tiny $\pm0.1135$}           \\
    UTopic        & 0.2382{\tiny $\pm0.0148$}          & 0.0618{\tiny $\pm0.0036$}           & 0.3197{\tiny $\pm0.0126$}           \\
    SCHOLAR       & 0.3470{\tiny $\pm0.0087$}          & 0.1585{\tiny $\pm0.0068$}           & 0.5126{\tiny $\pm0.0099$}           \\
    CLNTM         & 0.3508{\tiny $\pm0.0090$}          & 0.1596{\tiny $\pm0.0051$}           & 0.5082{\tiny $\pm0.0122$}           \\ \hline
    VICNTM        & {\bf 0.3521{\tiny $\pm0.0088$}} & \underline{0.1602}{\tiny $\pm0.0063$} & \underline{0.5143}{\tiny $\pm0.0116$} \\
    Deep VICNTM   & 0.3493{\tiny $\pm0.0055$}          & 0.1595{\tiny $\pm0.0057$}           & {\bf0.5162{\tiny $\pm0.0097$}}  \\
    VC-CLNTM      & 0.3459{\tiny $\pm0.0101$}          & 0.1591{\tiny $\pm0.0052$}           & \underline{0.5162}{\tiny $\pm0.0128$} \\
    Deep VC-CLNTM & 0.3468{\tiny $\pm0.0100$}          & 0.1587{\tiny $\pm0.0056$}           & \underline{0.5152}{\tiny $\pm0.0079$} \\
    VIC-CLNTM     & 0.3504{\tiny $\pm0.0094$}          & {\bf 0.1631{\tiny $\pm0.0057$}}  & \underline{0.5134}{\tiny $\pm0.0115$}           \\ \hline
    \end{tabular*}

\end{table*}

\begin{table*}[ht]
\centering
\caption{Results of NPMI of NTMs when $k=200$. Boldface indicates the best performance for each experiment. Underlining indicates superior performances of our models compared to the baselines for each experiment}
\label{tab:results200}
\begin{tabular*}{\textwidth}{@{\extracolsep\fill}lccc}
    \hline
    $k=200$              & 20NG                   & IMDb                   & Wiki                   \\ \hline
    ProdLDA       & 0.1925{\tiny $\pm0.0055$}          & 0.0727{\tiny $\pm0.0022$}          & 0.2064{\tiny $\pm0.0042$}          \\
    ECRTM         & 0.1811{\tiny $\pm0.0030$}          & 0.0550{\tiny $\pm0.0026$}          & 0.2567{\tiny $\pm0.0063$}          \\
    TSCTM         & 0.1745{\tiny $\pm0.0105$}          & 0.0775{\tiny $\pm0.0126$}          & 0.4006{\tiny $\pm0.0315$}          \\
    SCHOLAR       & 0.3173{\tiny $\pm0.0058$}          & 0.1288{\tiny $\pm0.0030$}          & 0.4557{\tiny $\pm0.0065$}          \\
    CLNTM         & 0.3187{\tiny $\pm0.0049$}          & 0.1292{\tiny $\pm0.0033$}          & 0.4555{\tiny $\pm0.0034$}          \\ \hline
    VICNTM        & {\bf 0.3212{\tiny $\pm0.0027$}} & {\bf 0.1297{\tiny $\pm0.0031$}} & {\bf 0.4571{\tiny $\pm0.0052$}}          \\
    Deep VICNTM   & \underline{0.3200{\tiny $\pm0.0051$}}          & 0.1284{\tiny $\pm0.0046$}          & 0.4553{\tiny $\pm0.0036$} \\
    VC-CLNTM      & \underline{0.3211{\tiny $\pm0.0050$}}          & \underline{0.1295}{\tiny $\pm0.0028$}    & 0.4534{\tiny $\pm0.0053$}          \\
    Deep VC-CLNTM & \underline{0.3199{\tiny $\pm0.0033$}}          & 0.1291{\tiny $\pm0.0035$}          & 0.4546{\tiny $\pm0.0050$}          \\
    VIC-CLNTM     & \underline{0.3202{\tiny $\pm0.0040$}}          & 0.1284{\tiny $\pm0.0019$}          & 0.4556{\tiny $\pm0.0031$}          \\ \hline
    \end{tabular*}
\end{table*}

\begin{table*}[ht]
\centering
\caption{Results of TD and IRBO of NTMs when $k=50$}
\label{tab:diversity}
{\fontsize{8}{10}\selectfont
    \begin{tabular*}{\textwidth}{@{\extracolsep\fill}lc@{\hspace{3pt}}c@{\hspace{3pt}}c@{\hspace{3pt}}c@{\hspace{3pt}}c@{\hspace{3pt}}c}
        \toprule
        & \multicolumn{2}{c}{20NG}                        & \multicolumn{2}{c}{IMDb}               & \multicolumn{2}{c}{Wiki}               \\ \midrule
        & TD        & IRBO                   & TD        & IRBO          & TD        & IRBO          \\ \midrule
ProdLDA & 0.8600{\tiny $\pm0.0105$}   & 0.9936{\tiny $\pm0.0008$} & 0.5252{\tiny $\pm0.0348$}   & 0.9454{\tiny $\pm0.0074$} & 0.8014{\tiny $\pm0.0332$}   & 0.9921{\tiny $\pm0.0020$} \\
ECRTM   & 0.9428{\tiny $\pm0.0486$}   & 0.9971{\tiny $\pm0.0030$} & 0.8812{\tiny $\pm0.0529$}   & 0.9922{\tiny $\pm0.0067$} & 0.9780{\tiny $\pm0.0093$}   & 0.9994{\tiny $\pm0.0003$} \\
TSCTM   & 0.8676{\tiny $\pm0.0422$}   & 0.9915{\tiny $\pm0.0051$} & 0.8229{\tiny $\pm0.0825$}   & 0.9958{\tiny $\pm0.0041$} & 0.9182{\tiny $\pm0.1223$}   & 0.9986{\tiny $\pm0.0021$} \\
UTopic  & 0.4940{\tiny $\pm0.0217$}   & 0.9567{\tiny $\pm0.0067$} & 0.1576{\tiny $\pm0.0131$}   & 0.7337{\tiny $\pm0.0278$} & 0.8362{\tiny $\pm0.0117$}   & 0.9941{\tiny $\pm0.0007$} \\
SCHOLAR & 0.9066{\tiny $\pm0.0055$}          & 0.9974{\tiny $\pm0.0005$}          & 0.8356{\tiny $\pm0.0253$}          & 0.9941{\tiny $\pm0.0013$} & 0.9894{\tiny $\pm0.0053$}          & 0.9997{\tiny $\pm0.0001$} \\
CLNTM   & 0.8990{\tiny $\pm0.0145$}          & 0.9973{\tiny $\pm0.0004$}          & 0.8448{\tiny $\pm0.0281$}          & 0.9948{\tiny $\pm0.0012$} & 0.9858{\tiny $\pm0.0100$}          & 0.9997{\tiny $\pm0.0002$} \\ \midrule
VICNTM  & 0.9000{\tiny $\pm0.0117$} & 0.9973{\tiny $\pm0.0004$} & 0.8448{\tiny $\pm0.0356$} & 0.9946{\tiny $\pm0.0018$} & 0.9858{\tiny $\pm0.0084$} & 0.9996{\tiny $\pm0.0002$} \\ \bottomrule
\end{tabular*}
}
\end{table*}

Tables~\ref{tab:results50} and~\ref{tab:results200} show the results in terms of NPMI when $k=50$ and $k=200$, respectively. 
When $k=50$, our primary model VICNTM achieved a higher topic coherence than the baselines (ProdLDA and SCHOLAR) and the state-of-the-art models (ECRTM and CLNTM) on all datasets and all of our variation models outperformed SCHOLAR and CLNTM on the Wiki dataset. 
On the IMDb dataset, one of our variation models (VIC-CLNTM) demonstrated significantly better performance. 
When $k=200$, VICNTM consistently achieved higher NPMI than all the baselines and the state-of-the-art models on all datasets\footnote{UTopic is not included, as its clustering method fails when $k$=200.}.
These results indicate that the VIC or VC regularization to latent topic representations or their high-dimensional embeddings improves the performance of NTMs.
Additionally, all the models built upon SCHOLAR showed significantly better performance than the others. This can likely be attributed to the annealing process implemented in SCHOLAR, which optimizes the model’s use of batch normalization during training when reconstructing BoW vectors.
For reference, the results in terms of perplexity can be found in Appendix~\ref{secA1}.

Table~\ref{tab:diversity} shows TD and IRBO when $k=50$. 
Compared to IRBO, TD shows more significant differences in the results.
Our primary model VICNTM achieved compareble or significantly better performance than most baselines, except for ECRTM, as the regularization in ECRTM is specifically designed to enhance topic diversity.
Note that TD measures the proportion of unique top words across topics.
Higher coherence in topics does not necessarily imply greater topic diversity due to the polysemous nature of words.

In summary, our experiments showed that our VICNTM worked more effectively in terms of topic coherence and was comparable or significantly better than the baselines in terms of topic diversity. 
The topic diversity performance may potentially change in accordance with the hyperparameters selected for the VICNTM.

\subsubsection{Qualitative Results}
\begin{table*}[ht]
\caption{Top-ten words in example topics learned from the datasets}\label{tab:topics}
{\fontsize{7}{6.5}\selectfont
\begin{tabular*}{\textwidth}{c@{}c@{}c@{}c}
    \toprule
    Dataset               & Method & NPMI   & \multicolumn{1}{l}{Top-ten words in a topic}                                                            \\ \midrule
    \multirow{2}{*}{20NG} & \multicolumn{1}{l}{CLNTM}  & 0.4288 & \multicolumn{1}{l}{disease patients study symptoms diet drug treatment medicine effects drugs}          \\
                          & \multicolumn{1}{l}{VICNTM} & 0.4949 & \multicolumn{1}{l}{patients diet disease symptoms treatment medicine patient medical effects doctor}    \\
    \multirow{2}{*}{IMDb} & \multicolumn{1}{l}{CLNTM}  & 0.1351 & \multicolumn{1}{l}{baseball concert bands album metal quaid band tag player sports}                     \\
                          & \multicolumn{1}{l}{VICNTM} & 0.2408 & \multicolumn{1}{l}{album concert musicians songs song band eddie bands singer musical}                  \\
    \multirow{2}{*}{Wiki} & \multicolumn{1}{l}{CLNTM}  & 0.2779 & \multicolumn{1}{l}{airline customer software customers fuselage airlines cockpit ceo airframe flights}  \\
                          & \multicolumn{1}{l}{VICNTM} & 0.4843 & \multicolumn{1}{l}{fuselage airframe cockpit airline flights takeoff airlines powerplant boeing runway} \\ \bottomrule
\end{tabular*}
}
\end{table*}

Table~\ref{tab:topics} lists example topics in the three datasets learned using our model and CLNTM to demonstrate that our model generates more coherent topics qualitatively. 
In the 20NG dataset, we observed that the NPMI of our model was higher than that of CLNTM, by which the word “study” tends to be commonly found in other topics. 
For the IMDb dataset, the example topic generated by our model focuses on music. 
In contrast, CLNTM’s topic includes both music and sports, such as “baseball”.
In the Wiki dataset, the example topic about “airplane” generated by CLNTM exhibits lower NPMI than that of our model. 
This was likely because the words “software” and “CEO” can also be commonly found in other topics. 
These findings suggest that our model is better suited for generating coherent topics than CLNTM.

\begin{figure*}[ht]
    \centering
    \includegraphics[width=1.0\textwidth]{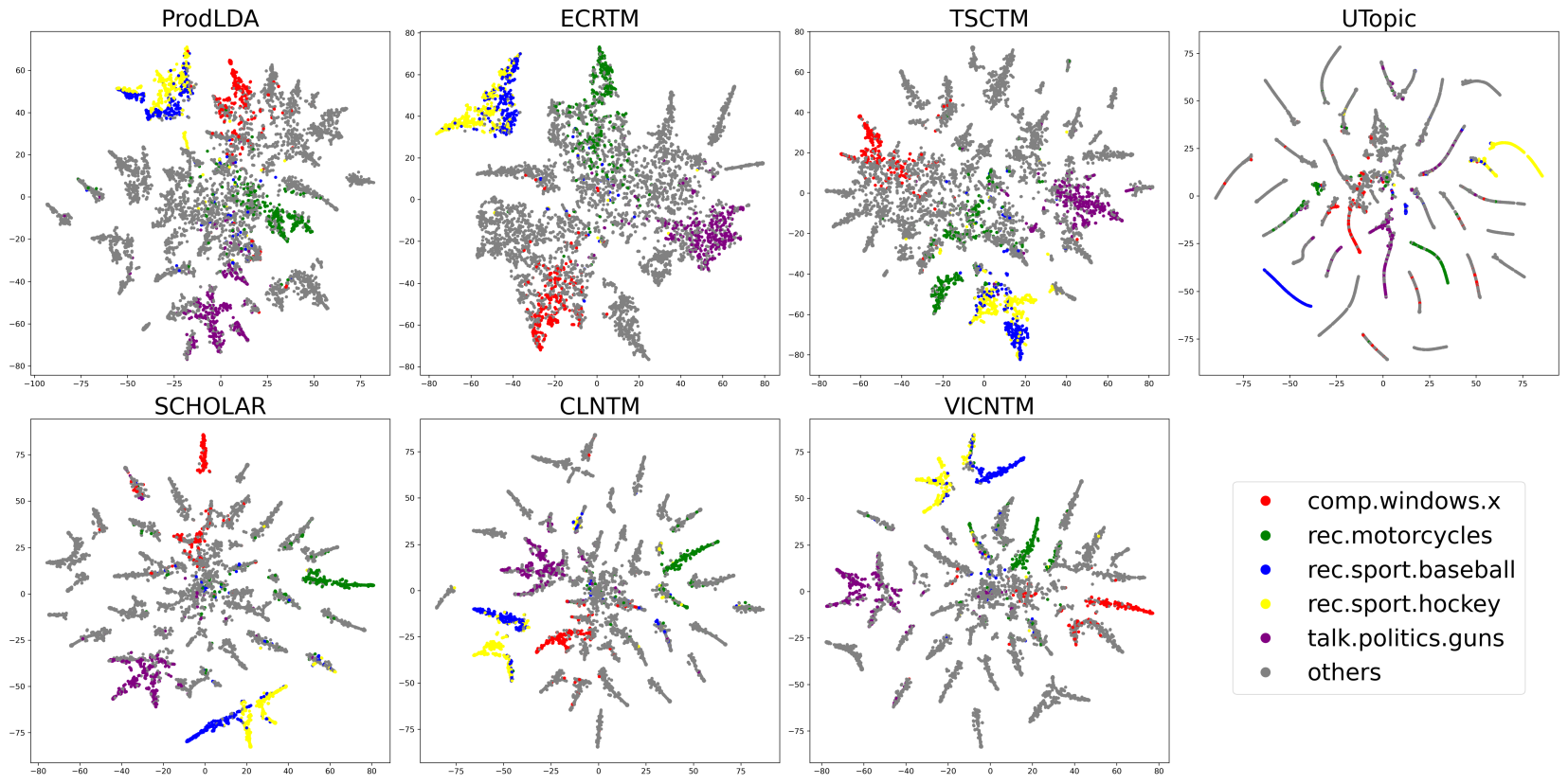}
    \caption{{\bf t-SNE plots of latent topic representations on the 20NG dataset}} \label{fig:tsne}
\end{figure*}
Fig.~\ref{fig:tsne} plots the latent topic representations generated from the 20NG dataset, by our primary model VICNTM, and all of the baselines, using the statistical method t-SNE for visualizing high-dimensional representations. 
We can see from Fig.~\ref{fig:tsne} that the representations generated by VICNTM exhibit, in general, a greater dispersion across the mapping space compared with those generated by CLNTM. 
This observation indicates that the incorporation of variance regularization in VICNTM assists in distinguishing between diverse documents and mitigating representation collapse. 
Note that VICNTM and CLNTM both use self-supervised learning. 
Nevertheless, VICNTM specifically addresses the collapse problem, whereas CLNTM does not.
Regarding SCHOLAR, the representations are similarly favorable. 
Unlike VICNTM and CLNTM, SCHOLAR does not rely on self-supervised learning. 
Consequently, instead of not having to consider the collapse problem, its performance on the NPMI metric is inferior to that of VICNTM.
The three SCHOLAR-based models (VICNTM, CLNTM, and SCHOLAR) exhibit tighter and more recongnizable clusters, while ProdLDA, ECRTM, and TSCTM show more scatterd and less coherent clustering. As for UTopic, the strip-like clusters are likely a result of the pre-clustering applied before training the NTM.

\subsection{Ablation study}\label{sec45}
\begin{table*}[ht]
\caption{Ablation study using 20NG data when $k=50$}\label{tab:ablation}
\begin{tabular*}{\textwidth}{@{\extracolsep\fill}lcc}
\toprule
Regularization & NPMI                   & Perplexity           \\ \midrule
Variance-Invariance                & 0.3494±0.0082          & 1678.7±17.4          \\
Invariance-Covariance              & 0.3505±0.0090          & 1694.3±27.6          \\
Invariance                         & 0.3489±0.0060          & \textbf{1667.1±19.7} \\ \midrule
VIC                                & \textbf{0.3521±0.0088} & 1685.3±36.9          \\ \bottomrule
\end{tabular*}
\end{table*}

We conducted an ablation study on the proposed model VICNTM to verify the effectiveness of the regularization terms. 
We kept the invariance regularization as it ensures that the model is learned in a self-supervised manner. 
The results in Table~\ref{tab:ablation} show that the covariance term has a greater impact than the variance term. 
In other words, the regularization of decorrelating different topics has a more significant contribution to topic quality than the regularization of maintaining the differences between different topic representations of documents. 
Overall, our proposed model VICNTM achieves the best performance in terms of NPMI when using all three regularizations.

\subsection{Experiments using adversarial data augmentation}\label{sec46}
\begin{figure*}[ht]
    \centering
    \includegraphics[width=1.0\textwidth]{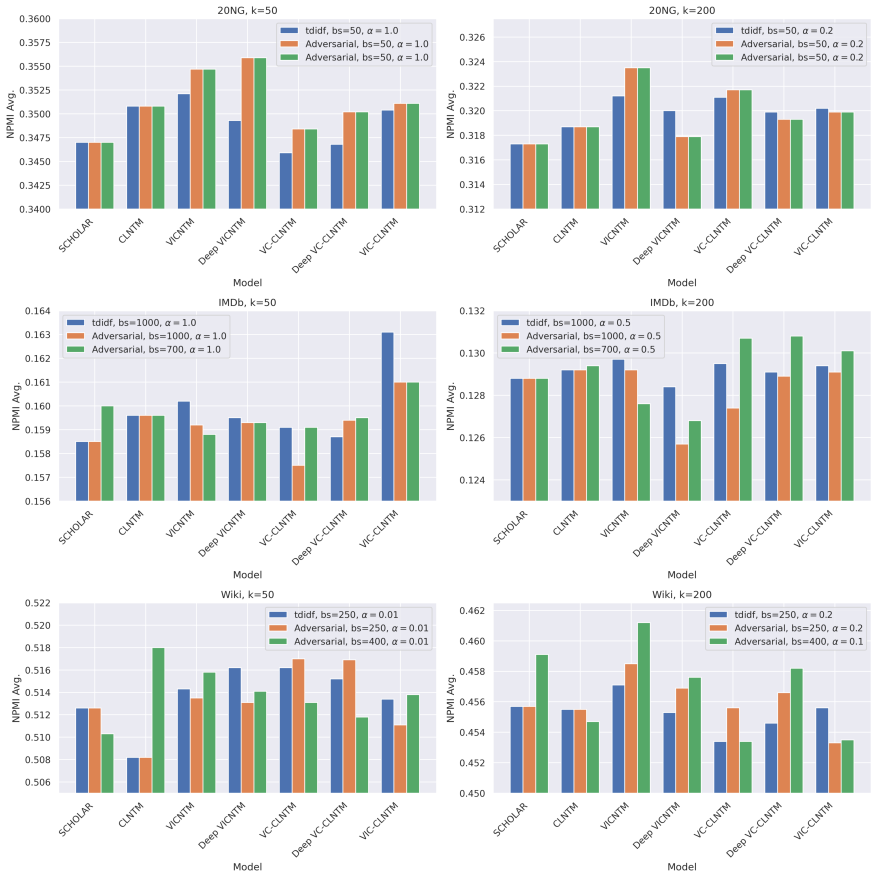}
    \caption{{\bf Results of NPMI when using different sampling strategies in our proposed models}} \label{fig:additional}
\end{figure*}

We now focus on the sampling methods described in Section~\ref{sampling strategies}, and evaluate how each sampling method works among the models discussed in the above sections.
As the heuristic tf-idf-based sampling method has the issue of generated samples becoming nearly identical to the true samples, which may lead to limited improvement, we also conducted the same experiments in two settings with the adversarial sampling strategy, aiming to bring potentially more improvement. 
In one setting, we used all the hyperparameters that were employed in the previous experimentation. In the other setting, we further optimized with the validation set for this additional experimentation, the most sensitive hyperparameters: batch size (bs) and logistic normal prior ($\alpha$).
We trained the sampling model with the three training datasets and averaged the outputs of the augmentation model over multiple epochs to obtain positive samples for each dataset. 
The positive samples are then used only in our proposed models. 

The results, including the initial experiments that used the tf-idf-based sampling strategy, are shown in Fig.~\ref{fig:additional}. 
Overall, despite the shift in the type of the best model among our proposed models when using the adversarial strategy, they still outperformed the baselines in the three different settings. 
In datasets with a broader range of topics (20NG and Wiki), our proposed models employing the adversarial strategy generally surpassed those using the tf-idf based strategy, as well as the baselines. 
In datasets with a sufficient number of documents (IMDb and Wiki), the models with the adversarial strategy and tuned hyperparameters demonstrated superior performance in most cases.

\section{Conclusions and discussions}
We proposed a self-supervised learning NTM with regularization to improve topic quality.
Inspired by the regularization-based self-supervised learning used in the image domain, we regularized the latent topic representations of anchor samples and positive samples with the VIC regularization when learning an NTM.
The regularization minimizes the differences between latent topic representations of similar documents, diversifies latent topic representations of different documents, and identifies individual topics within each latent topic representation. 
Through experiments on learning topics in large document collections, our proposed model produced more coherent topics than baselines and state-of-the-art models.
The results suggest that the VIC regularization imposed on the latent topic space contributes to better topic quality.
We also introduced a model-based adversarial sampling strategy, inspired by that developed for image classification~\cite{Suzuki_2022_CVPR}, to replace the heuristic tf-idf-based sampling strategy for generating positive samples aiming for potentially greater improvements. Both sampling strategies aided our proposed models in surpassing the comparison models.

For future work, we plan to further explore the incorporation of regularization in the high-dimensional space using an expander, similar to VICReg~\cite{bardes2022vicreg}.
We argue that the positive samples generated by replacing the tokens that have the lowest tf-idf scores do not offer sufficient diversity to notably enhance topic quality. 
Despite we have implemented a simple instance of the adversarial sampling strategy in this paper, we plan to consider a more sophisticated architecture and an augmentation model suited for text data augmentation.

\begin{appendices}
\section{Experimental results in terms of perplexity}\label{secA1}

\begin{table*}[ht]
    \centering
    \caption{Results of perplexity of NTMs when $k=50$. The boldface indicates the best performance for each experiment}
    \label{tab:perplexity50}
    \begin{tabular*}{\textwidth}{@{\extracolsep\fill}lccc}
        \hline
        $k=50$              & 20NG                   & IMDb                    & Wiki                    \\ \hline
        ProdLDA       & N/A                                & 4452.7{\tiny $\pm137.9$}           & 7986.8{\tiny $\pm912.3$}           \\
        ECRTM         & 3749.6{\tiny $\pm4236.9$}          & 4731.5{\tiny $\pm3501.3$}           & 5680.9{\tiny $\pm215.4$}           \\
        TSCTM         & {\bf 1554.0{\tiny $\pm72.3$}}          & 3183.7{\tiny $\pm626.1$}           & 6383.7{\tiny $\pm2417.9$}           \\
        UTopic        & N/A                              & N/A                               & N/A           \\
        SCHOLAR       & 1677.6{\tiny $\pm30.0$}          & 2242.3{\tiny $\pm18.5$}           & 3507.9{\tiny $\pm25.5$}           \\
        CLNTM         & 1790.6{\tiny $\pm36.1$}          & 2262.0{\tiny $\pm11.0$}           & 3488.0{\tiny $\pm44.0$}           \\ \hline
        VICNTM        & 1685.3{\tiny $\pm36.9$}          & {\bf 2156.4{\tiny $\pm11.0$}}           & 3531.6{\tiny $\pm25.4$} \\
        Deep VICNTM   & 1731.2{\tiny $\pm25.4$}          & 2167.3{\tiny $\pm15.9$}           & 3516.2{\tiny $\pm32.6$}  \\
        VC-CLNTM      & 1796.2{\tiny $\pm19.9$}          & 2257.1{\tiny $\pm13.7$}           & 3473.9{\tiny $\pm39.1$} \\
        Deep VC-CLNTM & 1776.2{\tiny $\pm43.4$}          & 2256.5{\tiny $\pm14.7$}           & {\bf 3467.4{\tiny $\pm32.3$}} \\
        VIC-CLNTM     & 1795.2{\tiny $\pm28.8$}          & 2254.6{\tiny $\pm14.6$}           & 3499.0{\tiny $\pm40.4$}           \\ \hline
        \end{tabular*}
    
\end{table*}

\begin{table*}[ht]
    \centering
    \caption{Results of perplexity of NTMs when $k=200$. The boldface indicates the best performance for each experiment}
    \label{tab:perplexity200}
    \begin{tabular*}{\textwidth}{@{\extracolsep\fill}lccc}
        \hline
        $k=200$              & 20NG                   & IMDb                   & Wiki                   \\ \hline
        ProdLDA       & 3459.7{\tiny $\pm173.0$}              & 7137.9{\tiny $\pm218.2$}          & 8742.4{\tiny $\pm385.2$}          \\
        ECRTM         & 67601.1{\tiny $\pm11544.1$}           & 23888.0{\tiny $\pm18172.8$}       & 6382.5{\tiny $\pm324.8$}          \\
        TSCTM         & 1733.9{\tiny $\pm209.2$}              & 2829.5{\tiny $\pm36.4$}          & 5398.9{\tiny $\pm251.9$}          \\
        SCHOLAR       & 1648.4{\tiny $\pm16.7$}               & 2512.0{\tiny $\pm46.5$}          & 3081.1{\tiny $\pm12.7$}          \\
        CLNTM         & 1759.6{\tiny $\pm163.4$}              & 2538.6{\tiny $\pm38.9$}          & 3056.6{\tiny $\pm16.7$}          \\ \hline
        VICNTM        & {\bf 1635.2{\tiny $\pm20.2$}}         & {\bf 2431.9{\tiny $\pm84.9$}}    & 3085.6{\tiny $\pm13.1$}          \\
        Deep VICNTM   & 1653.8{\tiny $\pm24.0$}               & 2439.1{\tiny $\pm49.2$}          & 3093.7{\tiny $\pm17.9$} \\
        VC-CLNTM      & 2031.4{\tiny $\pm932.2$}              & 2530.8{\tiny $\pm41.4$}          & 3064.3{\tiny $\pm25.0$}          \\
        Deep VC-CLNTM & 1731.7{\tiny $\pm33.1$}               & 2539.8{\tiny $\pm48.9$}          & 3058.8{\tiny $\pm17.4$}          \\
        VIC-CLNTM     & 1846.4{\tiny $\pm307.1$}              & 2537.1{\tiny $\pm56.9$}          & {\bf 3055.8{\tiny $\pm18.4$}}          \\ \hline
        \end{tabular*}
\end{table*}
    
Tables~\ref{tab:perplexity50} and~\ref{tab:perplexity200} show the experimental results in terms of perplexity when k=50 and k=200, respectively. 
Although a recent survey~\cite{wu2024survey} suggests that perplexity may not allow for fair comparisons due to differences in model architectures, overall, our proposed models achieved better performance in most cases. 
N/A in Table~\ref{tab:perplexity50} indicates that the model either produced an excessively large value or its architecture differs too significantly to compute perplexity.

\end{appendices}

\bibliographystyle{unsrt}
\bibliography{references}
\end{document}